\documentclass[]{fairmeta}
\usepackage{lineno}
\usepackage{calc}
\usepackage{amsmath}
\usepackage{caption}
\usepackage{multirow}
\usepackage{amsfonts}
\usepackage{xspace}
\usepackage[font=footnotesize]{subcaption}
\usepackage{booktabs}
\usepackage{pdfpages}
\usepackage{xcolor}
\usepackage{tabularx}
\usepackage{bbm}
\usepackage{graphicx}
\usepackage{algorithm}
\usepackage{algorithmic}
\usepackage{enumitem}
\usepackage{pifont}

\usepackage[export]{adjustbox}

\newcommand{\ours}{\textsc{Mixture-of-Minds}\xspace}

\newcommand{\tabref}[1]{Table~\ref{#1}}
\newcommand{\figref}[1]{Fig.~\ref{#1}}

\newcommand{\secref}[1]{\S\ref{#1}}
\newcommand{\appref}[1]{Appendix~\ref{#1}}

\title{Mixture-of-Minds: Multi-Agent Reinforcement Learning for Table Understanding}

\author{Yuhang Zhou}
\author{Mingrui Zhang}
\author{Ke Li}
\author{Mingyi Wang}
\author{Qiao Liu}
\author{Qifei Wang}
\author{Jiayi Liu}
\author{Fei Liu}
\author{Serena Li}
\author{Weiwei Li}
\author{Mingze Gao}
\author{Abhishek Kumar}
\author{Xiangjun Fan}
\author[\dagger]{Zhuokai Zhao}
\author[\dagger]{Lizhu Zhang}

\affiliation{Meta AI}

\contribution[\dagger]{Joint last author}

\abstract{Understanding and reasoning over tables is a critical capability for many real-world applications. 
Large language models (LLMs) have shown promise on this task, but current approaches remain limited.
Fine-tuning based methods strengthen language reasoning; yet they are prone to arithmetic errors and hallucination. 
In contrast, tool-based methods enable precise table manipulation but rely on rigid schemas and lack semantic understanding.
These complementary drawbacks highlight the need for approaches that integrate robust reasoning with reliable table processing.
In this work, we propose \ours, a multi-agent framework that decomposes table reasoning into three specialized roles: planning, coding, and answering.
This design enables each agent to focus on a specific aspect of the task while leveraging code execution for precise table manipulation. 
Building on this workflow, we introduce a self-improvement training framework that employs Monte Carlo Tree Search (MCTS) rollouts to generate pseudo-gold trajectories and optimize agents with reinforcement learning (RL). 
Extensive experiments show that \ours delivers substantial gains, reaching 62.13\% on TableBench and surpassing OpenAI-o4-mini-high. 
These results demonstrate the promise of combining structured multi-agent workflows with RL to advance table understanding.
}

\date{\today}
\correspondence{Yuhang Zhou, Zhuokai Zhao and Lizhu Zhang at \email{\{zyhang, zhuokai, lizhu\}@meta.com}\\}

\begin{document}

\maketitle

\section{Introduction}
\label{sec:intro}

Understanding and reasoning over tables is a fundamental capability for many real-world applications, including finance, healthcare, and knowledge management~\citep{zhang2025surveytable, cheng2025survey, fang2024large}. 
Compared with free text, tables present unique challenges: they are dense, structured, and often contain noisy or incomplete entries. 
Effective table understanding therefore requires models to retrieve relevant cells, perform reasoning steps such as arithmetic or comparison, and map the results into natural language answers~\cite{wang2024chain, yang2025table}.

Large language models (LLMs) have recently become the backbone of state-of-the-art table understanding approaches~\cite{sui2024table, tang2025llm}. 
Prior work generally follows two directions. 
One seeks to enhance the intrinsic reasoning capability of LLMs through model-only training, using techniques such as supervised finetuning (SFT) or reinforcement learning (RL)~\cite{wu2025table}. 
The other direction leverages external tools, most commonly Python or SQL, to transform tables, filter noise, and extract useful information before the LLM generates the final answer~\cite{sun2023sql, shi2024survey, gao2023text}. 
%
%
The first stream provides flexible language-based reasoning but is prone to arithmetic errors~\citep{bertolazzi2025validation}, structural confusion~\citep{orgad2024llms}, and hallucination~\citep{kalai2025language}, while the second enables precise table manipulation but relies on rigid schema and lacks semantic depth or intent understanding~\citep{rath2025structured}.

To address these challenges, we propose \textbf{\ours}, a multi-agent framework that integrates the advantages of both directions. 
Instead of relying on a single agent to solve the task end-to-end, \ours decomposes table reasoning into three specialized roles: a \emph{planning agent} that outlines reasoning steps, a \emph{coding agent} that generates and executes code to transform the table, and an \emph{answering agent} that derives the final answer from structured evidence. 
This decomposition not only makes the workflow more interpretable but also allows each LLM agent to focus on a well-defined part of the problem.
%


However, training such a multi-agent system is significantly more challenging than training a single agent. 
The key difficulty is the lack of intermediate supervision: while final answers can be supervised, there are usually no gold-standard plans or code traces to guide the planning and coding agents. 
To address this, we introduce a self-improvement training framework that leverages Monte Carlo Tree Search (MCTS)-style rollouts~\cite{browne2012survey} to automatically collect high-quality intermediate trajectories as pseudo-gold supervision. 
MCTS explores many alternative reasoning paths and keeps only those that lead to a correct final answer.
By assuming that the intermediate steps along successful paths are likely to be valid, we turn the hard problem of supervising reasoning into a sequence of verifiable subtasks.
%
%
%
This enables the use of RL methods, such as Group Relative Policy Optimization (GRPO)~\citep{shao2024deepseekmath}, with specifically designed reward functions that jointly capture plan quality, code execution validity, and final answer accuracy. 

Our contributions and findings can be summarized as follows:  
\begin{itemize}
    \item[\ding{68}] We propose a novel agent workflow for table understanding that decomposes question solving into planning, coding, and answering, with specialized agents for each stage.  
    \item[\ding{68}] We propose a self-improvement training framework that leverages MCTS-based data generation to yield verifiable intermediate supervision for multi-agent optimization.
    \item[\ding{68}] By integrating both the workflow and training paradigm, we introduce \ours, which achieves substantial improvements over strong baselines. Extensive experiments show that \ours attains 62.13\% accuracy on TableBench~\cite{wu2025tablebench} with smaller LLMs and surpasses state-of-the-art models such as OpenAI o4-mini-high~\citep{jaech2024openai}.
\end{itemize}

\section{Related Work}
\label{sec:related}

\subsection{Tabular Data Understanding}

Table Question Answering (Table QA) has advanced alongside the development of increasingly complex evaluation benchmarks~\cite{mueller2019answering, jin2022survey, wang2024chain}. 
%
%
Early work focused on simple fact checking~\cite{pasupat2015compositional, iyyer2017search, chen2019tabfact}; later efforts incorporated tasks requiring complex mathematical reasoning \cite{chen2021finqa, katsis2021ait}; and recent benchmarks extend to multimodal settings that integrate tables with other modalities~\cite{talmor2021multimodalqa}.
This evolution has expanded table understanding tasks from single-cell extraction to broader reasoning and calculations over entire table contexts.

To meet the need for reasoning over entire tables, prior work has generally followed two directions: (1) enhancing the reasoning ability of LLMs through model-only inference, using table-specific RL or SFT methods without external tools~\citep{yang2025table, wu2025table, sui2024table, lei2025reasoning}; and (2) leveraging external tools such as Python or SQL to transform tables, filter noise, and extract useful information before feeding the processed data back into LLMs~\cite{zhang2024reactable, wang2024chain, ye2023large, zhang2024alter}.
Compared with these approaches, our work integrates the advantages of both streams. 
%
%
We propose a novel agent workflow that first decomposes table reasoning into planning, coding, and answering, and introduce an MCTS-based training framework that systematically improves the agents under this design.

\subsection{Agentic Reinforcement Learning}

Agentic Reinforcement Learning (Agentic RL) views LLMs as learnable policies within sequential decision-making loops, where RL equips them with agentic capabilities for long-horizon reasoning and interaction in dynamic environments~\citep{zhang2025landscapeagenticreinforcementlearning}. 
Using methods such as Proximal Policy Optimization (PPO)~\citep{schulman2017proximal} or GRPO~\citep{shao2024deepseekmath} to fine-tune agent LLMs, researchers have enhanced their planning, reasoning, and tool-use abilities~\citep{lu2025pilotrl, schick2023toolformer, wu2025vtool, guo2025deepseek}. 
These advances have led to powerful agents applicable across domains ranging from search, coding to mathematics~\cite{feng2025retool, li2024acecoder, jin2025search}. 
In the field of tabular data understanding, most prior work has trained either a single agent end to end or multiple agents jointed in a single-round RL setup~\citep{liu2025skyrlsql, chen2025breaking}.
In contrast, we decompose table reasoning into specialized sub-agents and introduce an MCTS-based data generation and training pipeline that optimizes each agent through targeted RL, enabling more effective self-improvement.
%
%

\section{\ours Agents Workflow}
\label{sec:workflow}

%
%

Tables typically contain far more information than any single question requires, with many irrelevant entries that make it difficult for LLMs to locate useful facts and perform the necessary multi-hop reasoning.
In such settings, relying on a single inference step to both plan, generate code, and produce the final answer often leads to incomplete reasoning, execution errors, or misaligned outputs.

\begin{figure}
    \centering
    \includegraphics[width=0.85\linewidth]{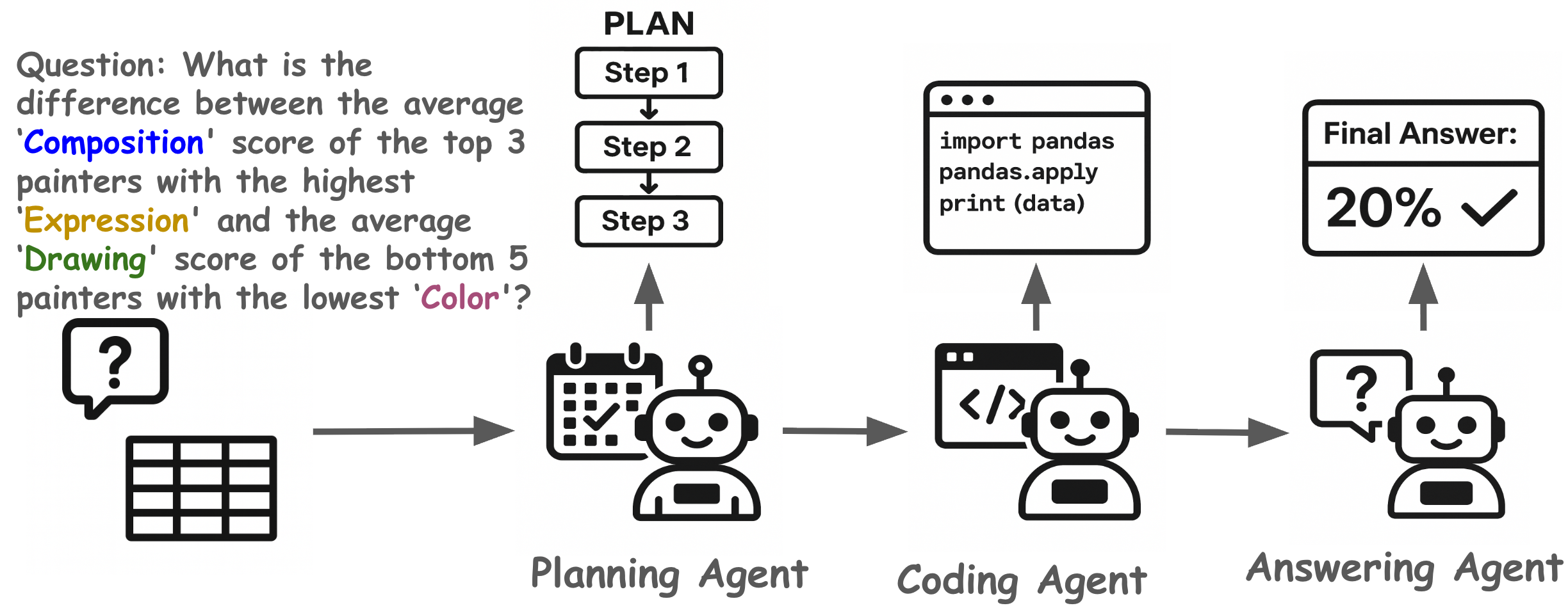}
    \caption{
    Overview of the \ours workflow. 
    The process is decomposed into three agents: (1) the \textit{planning agent}, which outlines a structured plan; (2) the \textit{coding agent}, which generates and executes code; and (3) the \textit{answering agent}, which synthesizes the plan, table, and code outputs to produce the final answer.
    %
    }
    \label{fig:agent_workflow}
\end{figure}

To address these challenges, we propose \ours, an agent workflow that leverages code as an external tool for precise table processing and decomposes the problem into three coordinated stages: \emph{planning}, \emph{coding}, and \emph{answering}. 
As shown in \figref{fig:agent_workflow}, the planning agent first outlines a reasoning plan that guides the coding agent toward more accurate program generation.
The coding agent then executes the program to transform the table into an intermediate representation, which serves as structured evidence. 
Finally, the answering agent combines this evidence with the question and the plan, relying on the model's reasoning and instruction-following abilities to produce the final answer.
Each agent is guided by a specialized prompt, and together they form a modular and interpretable pipeline. 
Detailed prompt templates for each agent are provided in \appref{sec:prompts}.

Formally, let $q$ denote a question and $T$ its associated structured table.
Let $\pi_\theta$ be the LLM-based agent parameterized by $\theta$.
Given a question–table pair $(q, T)$, the workflow proceeds as follows:

\paragraph{Planning agent.}  
The planning agent $\pi^{\text{plan}}_\theta$ generates a plan $p$ wrapped in $\langle \texttt{plan} \rangle \cdot \langle/\texttt{plan}\rangle$ tags. 
The plan provides structured reasoning instructions that guide subsequent stages, helping prevent unguided and potentially ineffective exploration.

\paragraph{Coding agent.}  
Conditioned on $(q, T, p)$, the coding agent $\pi^{\text{code}}_\theta$ outputs executable code $c$, wrapped in Python markdown delimiters. 
Executing $c$ yields an ordered sequence of intermediate table representations $\textsc{T}'=(T'_1, T'_2,\dots,T'_k)$, similar to~\citep{wang2024chain}.
These intermediate results provide verifiable operations and reduce hallucinations common in free-form reasoning.

\paragraph{Answering agent.}  
The answering agent $\pi^{\text{ans}}_\theta$ produces the answer $\hat{y}$, formatted within $\langle \texttt{answer} \rangle \cdot \langle/\texttt{answer}\rangle$ tags. 
By conditioning on $(q, p, \textsc{T}')$, the agent grounds its response in both the original query and the structured evidence.

This decomposition follows the natural workflow of table reasoning: planning the reasoning steps, deriving intermediate sub-results, and integrating them into a final answer.
By explicitly separating these stages, the pipeline improves interpretability, allows error diagnosis at intermediate steps, and makes the reasoning process more robust to failures in any single component.  

\paragraph{Improved test-time scaling support.}
Prior work has shown that allocating more test-time compute, such as sampling multiple trajectories or extending their length, can improve reliability and accuracy~\citep{muennighoff2025s1, zhang2025survey}.
Unlike direct inference, which produces an answer in a single step, our workflow with task decomposition naturally increases token usage. 
This modular design and structure further makes our approach well suited to test-time scaling (TTS). 
Because each agent serves a distinct role, we can adjust decoding strategies, such as sampling more rollouts during early planning and coding to encourage diverse reasoning paths, and fewer rollouts during later answering to ensure stability. 
This fine-grained control enables early branching, a strategy known to enhance TTS~\citep{yang2025alignment, fu2025deep}, but difficult to realize in single-model setups, where one must repeat all chain-of-thought (CoT) traces or cannot easily determine which tokens require more exploration.

In the next section, we describe how we build a specifically-designed training framework to further enhance \ours.

\section{\ours Agents Training}
\label{sec:agent_train}

\begin{figure*}[t]
    \centering
    \includegraphics[width=\linewidth]{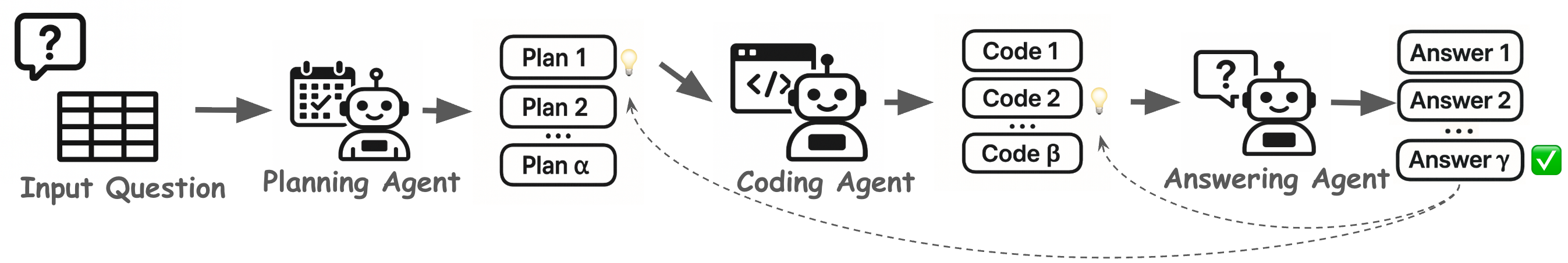}
    \caption{
        Illustration of the \ours workflow with MCTS-style rollouts.
        Each agent produces multiple candidate inferences and we evaluate the final outputs to identify the correct answer. 
        In this example, Answer~$\gamma$ is judged as the correct answer, and we trace back to Code~2 and Plan~1—the steps that produced it—which are then marked as pseudo-gold intermediate supervision for training.
    }
    \label{fig:mcts_diagram}
\end{figure*}

With the design of our \ours agent workflow established, the next challenge is to iteratively improve the models within this pipeline.
We adopt a \textbf{step-by-step training strategy}, where the planning, coding, and answering agents are optimized sequentially and collaboratively.
To provide reliable supervision at each stage, we employ MCTS–style rollouts~\citep{browne2012survey} to generate diverse trajectories and extract high-quality intermediate plans and codes as gold-standard supervision (see \figref{fig:mcts_diagram}). 
Each agent is then refined using GRPO~\citep{shao2024deepseekmath} with structured and correctness-driven rewards. 
This progressive training ensures that agents improve in their designated roles while the overall workflow converges toward robust table reasoning. 
%


\subsection{MCTS-style Data Generation}
\label{section:mcts}
To effectively finetune the planning, coding, and answering agents within the \ours workflow, it is necessary to obtain gold-standard intermediate supervision, namely reference-quality plans and codes (we use the existing final answer as the label for the answering agent). 
Since such annotations are not available in existing datasets, we introduce an MCTS-style rollout procedure to automatically construct high-quality intermediate trajectories.
Formally, following notations in \secref{sec:workflow},
given a question-table pair $(q, T)$, the process unfolds in three stages:

\paragraph{Plan generation.}  
The planning agent $\pi^{\text{plan}}_\theta$ generates $\alpha$ candidate plans with temperature $\tau_p$:
$\{p_1, p_2, \dots, p_\alpha\}.$

\paragraph{Code generation.}  
For each plan $p_i$, the coding agent $\pi^{\text{code}}_\theta$ produces $\beta$ executable code candidates with temperature $\tau_c$:
$
\{c_{i,1}, c_{i,2}, \dots, c_{i,\beta}\}.
$

\paragraph{Answer generation.}  
Each code candidate $c_{i,j}$ is executed to obtain intermediate table representations $\textsc{T}'_{i,j}$. 
For every $(q, p_i, \textsc{T}'_{i,j})$ tuple, the answering agent $\pi^{\text{ans}}_\theta$ generates $\gamma$ candidate answers with temperature $\tau_a$:
$
\{\hat{y}_{i,j,1}, \hat{y}_{i,j,2}, \dots, \hat{y}_{i,j,\gamma}\}.
$

A rollout trajectory is considered \textit{successful} if at least one $\hat{y}_{i,j,k} \in \mathcal{A}(p_i, c_{i,j})$ matches the ground-truth answer $y^\star$. 
In such cases, both the intermediate plan $p_i$ and code $c_{i,j}$ are considered as high-quality supervision signals.
Therefore, the generated dataset $\mathcal{D}$ is defined as:

\scalebox{0.9}{%
\begin{minipage}{\linewidth}
\centering
\begin{equation*}
    \mathcal{D} = \{(q, T, p_i, c_{i,j}, \hat{y}_{i,j,k}) \mid 
\exists (i,j,k)
\ \text{s.t. } \hat{y}_{i,j,k} = y^\star \}
\end{equation*}
\vspace{0.01in}
\end{minipage}
}

\noindent which is used for self-improvement training of different agents. 
This procedure provides an automatic mechanism for collecting reliable intermediate annotations, thereby enabling stepwise optimization of the entire workflow.

\subsection{
Agent Training with GRPO
}
\label{section:grpo}

To optimize the agents in our workflow, we adopt GRPO~\citep{shao2024deepseekmath}, a RL algorithm that refines language model policies using relative reward signals within a group of responses.
Specifically, GRPO refines a policy $\pi_{\phi}$ by sampling multiple candidate outputs for each input and updating the policy based on their relative quality. 
Formally, given a prompt $x$ (e.g., $x=(q, T)$ for planning agent) from dataset $\mathcal{D}$, the old policy $\pi_{\phi_{\text{old}}}(y|x)$ samples $G$ responses
$
\textsc{Y}(x) = \{y_1, y_2, \dots, y_G\},
$
where each response $y_i \in \textsc{Y}(x)$ is assigned a scalar reward $r(x, y_i)$ (defined differently for planning, coding, and answering agents as detailed in the later subsections). 
The group-normalized advantage is then computed as:
$A(x,y_i) = \frac{r(x,y_i) - \bar{r}(x)}{\sigma_r(x)},$
where 
$\bar{r}(x)$ and $\sigma_r(x)$ are the mean and standard deviation of rewards within the group. 
The policy is then updated as:

\begin{equation*} 
\label{eq:grpo_objective_multline}
\mathcal{J}_{\text{GRPO}}(\theta) = \mathbb{E}_{q, \{o_i\}} \Biggl[ \frac{1}{G} \sum_{i=1}^{G} \min \Biggl( \frac{\pi_{\phi}(y_i|x)}{\pi_{\phi_{\text{old}}}(y_i|x)} A_i, \text{clip}\left(\frac{\pi_{\phi}(y_i|x)}{\pi_{\phi_{\text{old}}}(y_i|x)}, 1-\epsilon, 1+\epsilon\right) A_i \Biggr) \Biggr]  - \beta \, \mathbb{D}_{\text{KL}}[\pi_\theta \| \pi_{\text{ref}}],
\end{equation*}

\noindent where $\epsilon$ is the clipping parameter and $\beta$ controls the KL penalty against a reference policy $\pi_{\text{ref}}$.

In the context of our training framework, $x$ denotes the input prompt for each agent as described in \secref{sec:workflow}, while the candidate responses $y_i$ correspond to either sampled plans, codes, or answers depending on the agent being optimized.

\paragraph{Planning agent training.}
\label{sec:plan_agent}
%
For the planning agent, we first apply the MCTS-style rollout procedure (\secref{section:mcts}) with hyperparameters $(\alpha_{\text{plan}}, \beta_{\text{plan}}, \gamma_{\text{plan}})$ to generate trajectories for each $(q,T)$.
%
We retain only successful traces where the final answer matches $y^\star$, and use the extracted plans $\{p^\star\}$ as gold-standard supervision.

During GRPO training, the planning agent $\pi^{\text{plan}}_\theta$ generates $G$ candidate plans in each group.
Each plan $p_i$ receives a reward
$
r^{\text{plan}}(p_i) = 0.1\, r^{\text{fmt}}(p_i) + 0.9\, r^{\text{BLEU}}(p_i, p^\star),
$
where $r^{\text{fmt}}$ checks correct formatting and $r^{\text{BLEU}}$ is the BLEU score \cite{papineni2002bleu} against the gold-standard plan. 

\paragraph{Coding agent training.}
\label{sec:code_agent}
For coding, we leverage the fine-tuned planning agent to provide candidate plans, while the coding and answering agents remain base LLMs. 
MCTS rollouts with $(\alpha_{\text{code}}, \beta_{\text{code}}, \gamma_{\text{code}})$ yield trajectories, from which we keep only successful traces and extract gold-standard codes $\{c^\star\}$, using $(q, T, p^\star)$ as the corresponding training input.

In GRPO training, each candidate code $c_i$ in the group with group size $G$ is scored with a mixed reward:
$r^{\text{code}}(c_i) = 0.1\, r^{\text{fmt}}(c_i) + 0.2\, r^{\text{exec}}(c_i) + 0.2\, r^{\text{op}}(c_i) + 0.5\, r^{\text{out}}(c_i, c^\star),$
where $r^{\text{fmt}}$ checks Python markdown formatting, $r^{\text{exec}}$ indicates successful execution, $r^{\text{op}}$ measures code structural similarity to $c^\star$ (via Pandas-operation F1), and $r^{\text{out}}(c_i, c^\star)$ measures the alignment between the generated code output and the gold-standard code output, computed using BLEU score. 
We provide more details of the reward design in \appref{sec:reward}.

\paragraph{Answering agent training.}
For the answering agent, both the planning and coding agents are fine-tuned, while the answering agent remains base. 
MCTS rollouts with $(\alpha_{\text{ans}}, \beta_{\text{ans}}, \gamma_{\text{ans}})$ generate trajectories, from which we keep correct answers $\{\hat{y}_i \mid \hat{y}_i = y^\star\}$ as supervision, using $(q, p^\star, \textsc{T}'^\star)$ as the corresponding input prompt. 

During GRPO training, each candidate answer $\hat{y}_i$ in the group with group size $G$ is scored as
$
r^{\text{ans}}(\hat{y}_i) = 0.1\, r^{\text{fmt}}(\hat{y}_i) + 0.9\, r^{\text{EM}}(\hat{y}_i),
$
where $r^{\text{fmt}}$ checks special-tag formatting and $r^{\text{EM}}$ denotes exact match with $y^\star$.

Together, these three stages form a step-wise training pipeline: the planning agent is first optimized to produce reliable plans, the coding agent is then trained using these plans as inputs, and finally the answering agent is refined with both plans and codes. 
This progressive strategy ensures that each agent benefits from the supervision generated by the previously fine-tuned components, leading to a consistent and robust end-to-end workflow.

\section{Experiments}
\label{sec:experiments}

We design our experiments to rigorously evaluate the effectiveness of both the \ours agents workflow and training framework. 

\subsection{Experimental Setup}
\paragraph{Dataset.}
%
For training, we adopt TableInstruct~\citep{wu2025tablebench}, which contains 4,897 unique question–answer pairs for table understanding.
Details of the data pre-processing are provided in \appref{sec:data_processing}. 
For in-domain evaluation, we use TableBench~\cite{wu2025tablebench}, focusing on three representative datasets: Fact Checking, Numerical Reasoning, and Data Analysis, comprising a total of 836 test questions. 
To further assess generalization, we also evaluate on FinQA~\cite{chen2022finqadatasetnumericalreasoning}, which includes 1,147 financial contexts, questions, and tables, serving as an out-of-domain benchmark.

\paragraph{Baselines.}
We compare our framework against three categories of baselines: (1) Strong proprietary and open-source LLMs that are several orders of magnitude larger than ours, including OpenAI GPT-5, o4-mini-high~\citep{jaech2024openai}, o4-mini, o3-mini, Gemini-2.5-Pro~\citep{huang2025gemini, comanici2025gemini}, Grok-4~\citep{grok4}, Claude-4-Sonnet~\citep{Claude-4}, and DeepSeek-R1~\citep{guo2025deepseek};
(2) Alternative training methodologies: we further consider models initialized from the same base LLMs but trained with different strategies, including GRPO and Decoupled Clip and Dynamic sampling Policy Optimization (DAPO)~\cite{shao2024deepseekmath, yu2025dapo};
and (3) state-of-the-art agent frameworks with integrated training recipes tailored for table understanding, such as Table-LLM~\citep{zhang2024tablellm} and Table-R1~\citep{wu2025table}.

\paragraph{Models.}
We adopt a diverse set of LLMs with different architectures and sizes. 
Specifically, we use LLaMA-3.1-8B, LLaMA-3.3-70B~\citep{grattafiori2024llama}, Qwen-3-8B, Qwen-3-32B~\citep{yang2025qwen3}, Llama-3\_3-Nemotron-Super-49B-v1\_5~\citep{bercovich2025llamanemotronefficientreasoningmodels} and Gemma-3-27B~\citep{team2025gemma}. 
This selection covers a wide range of LLM families (LLaMA, Qwen, and Nemotron) and sizes (8B-70B), providing a robust testbed for evaluating our agent workflow and training recipe.

\paragraph{Implementation details.} 
The numbers of sampled plans ($\alpha$), codes ($\beta$), and answers ($\gamma$) during MCTS rollouts differ by agent, as summarized in \tabref{tab:mcts_hparams} in \appref{sec:hyper}. 
%
%
The temperature ($\tau_p, \tau_c , \tau_a $) of MCTS and GRPO rollouts is fixed at $1.0$. 
Other training details can be found in \appref{sec:hyper}.

\subsection{Results}
\label{subsec:result}

%
We begin by verifying that the \ours agents workflow itself provides clear benefits over standard model-only inference. 
We then assess additional improvement gained through our \ours training framework. 
%


\paragraph{\ours agent workflow significantly outperforms direct inference.}  
We first verify the effectiveness of the \ours pipeline compared to direct model-only inference on table understanding tasks. 
In this setting, the same base LLMs are evaluated under two conditions: (1) standard inference, where the model directly predicts the answer given the question and table with the CoT~\citep{wei2022chain} prompts; 
and (2) the \ours agents workflow, where the model operates through the planning, coding, and answering agents \textit{without} training.
This comparison isolates the impact of the workflow design itself, independent of our training framework. 
%
%

\begin{table}[t]
\setlength{\tabcolsep}{16pt}
\centering
\small
\begin{tabular}{lccccc}
\toprule
Model & Setting & FC & NR & DA & Avg. \\
\midrule
\multirow{2}{*}{LLaMA-3.1-8B} 
 & Direct Inference & \textbf{54.17} & 7.30 & 16.20 & \textbf{15.42} \\
 & \ours & 30.21 & \textbf{8.08} & \textbf{19.77} & 14.58 \\
\midrule
\multirow{2}{*}{Qwen3-32B} 
 & Direct Inference & 81.25 & 72.54 & 20.06 & 49.02 \\
 & \ours & \textbf{81.25} & \textbf{73.99} & \textbf{20.06} & \textbf{49.68} \\
\midrule
\multirow{2}{*}{Qwen3-8B} 
 & Direct Inference & 78.12 & 53.90 & \textbf{24.11} & 41.98 \\
 & \ours & \textbf{79.17} & \textbf{68.18} & 21.37 & \textbf{47.38} \\
\midrule
\multirow{2}{*}{Gemma3-27B} 
 & Direct Inference & \textbf{79.17} & 54.66 & 24.73 & 42.67 \\
 & \ours & 78.12 & \textbf{61.11} & \textbf{30.93} & \textbf{47.83} \\
\midrule
\multirow{2}{*}{Nemotron-49B} 
 & Direct Inference & 76.04 & 58.69 & \textbf{33.48} & 44.42 \\
 & \ours & \textbf{76.04} & \textbf{71.28} & 26.13 & \textbf{50.30} \\
\midrule
\multirow{2}{*}{LLaMA-3.3-70B} 
 & Direct Inference & 72.16 & 33.61 & \textbf{34.53} & 36.25 \\
 & \ours & \textbf{78.12} & \textbf{69.52} & 34.23 & \textbf{52.88} \\
\bottomrule
\end{tabular}
\caption{
    Comparison between direct inference and \ours agent workflow on Fact Checking (FC), Numerical Reasoning (NR), and Data Analysis (DA) tasks~\citep{wu2025tablebench}. 
}
\label{tab:ours_vs_direct}
\end{table}
From \tabref{tab:ours_vs_direct}, we observe that the \ours pipeline consistently improves performance over direct inference. 
For strong models such as LLaMA-3.3-70B and Nemotron-49B, the average improves by +16.6\% and +5.9\%, respectively, driven primarily by large gains in numerical reasoning. 
Qwen3-8B and Gemma3-27B also show meaningful improvements of +5.4\% and +5.2\%, confirming that even mid-sized models benefit from structured agent reasoning. 
While LLaMA-3.1-8B remains roughly unchanged overall, it still benefits in numerical reasoning and data analysis.

These results demonstrate the effectiveness of the \ours workflow: decomposing table reasoning into planning, coding, and answering stages yields a substantial improvement compared to direct model predictions, with a particularly strong impact on numerical reasoning tasks where structured intermediate steps are essential.
%


\paragraph{The \ours training recipe further boosts performance, surpassing LLMs several orders of magnitude larger.}
Next, we evaluate the effectiveness of \ours training recipe.
We report in-domain (Fact Checking, Numerical Reasoning and Data Analysis tasks~\citep{wu2025tablebench}) and out-of-domain (FinQA~\citep{chen2022finqadatasetnumericalreasoning}) results comparing our methods and the baselines in \tabref{tab:ours_training} and \tabref{tab:ours_training_finqa}, respectively.
%


\begin{table*}[t]
\setlength{\tabcolsep}{12pt}

\centering
\small
\begin{tabular}{l l cccc}
\toprule
Model & Method & FC & NR & DA & Average \\
\midrule
OpenAI-o4-mini-high & Direct Inference & 82.29 & 81.11 & 42.30 & 61.69 \\
OpenAI-o4-mini & Direct Inference &83.33	&80.35	&40.5	&60.75   \\
GPT-5 & Direct Inference & 83.33 & 82.37 & 36.04 & 59.94 \\
OpenAI-o3-mini & Direct Inference & 86.46 & 82.07 & 35.56 & 59.90 \\
Gemini-2.5-Pro &Direct Inference &84.38	&79.6	&31.86 &57.18 \\
DeepSeek-R1 & Direct Inference & 82.29 & 75.51 & 35.06 & 56.31 \\
Claude-4-Sonnet &Direct Inference &81.25 &75.57	&31.14	&54.75  \\
Grok-4 & Direct Inference & 83.33 & 73.80 & 40.53 & 57.80 \\
\midrule
\multirow{6}{*}{Qwen3-8B} 
 & Direct GRPO & 82.29 & 78.34 & 24.98 & 53.69 \\
 & Direct DAPO & 81.25 & 75.31 & 25.84 & 52.58 \\
 & Table-R1 & -- & -- & -- & 49.30 \\
 & Table-LLM & -- & -- & -- & 35.88 \\
 & \ours   & 84.38 & 74.75 & 38.29 & 57.44 \\
 & \ours (w/ TTS) & \textbf{86.46} & \textbf{77.83} & \textbf{41.59}  & \textbf{60.35} \\
\midrule
\multirow{6}{*}{LLaMA-3.1-8B} 
 & Direct GRPO & 81.25 & 48.87 & 18.49 & 37.86 \\
 & Direct DAPO & 77.08 & 43.83 & 15.51 & 34.00 \\
 & Table-R1 & -- & -- & -- & 42.78 \\
 & Table-LLM & -- & -- & -- & 30.77 \\
 & \ours   & 70.83 & 59.70 & 31.69 & 46.72 \\
 & \ours (w/ TTS) & \textbf{81.25} & \textbf{67.51} & \textbf{36.80}  & \textbf{53.31} \\
\midrule
\multirow{4}{*}{Gemma3-27B} 
 & Direct GRPO & \textbf{84.38} & 67.51 & 23.14 & 48.36 \\
 & Direct DAPO & 80.21 & 65.74 & 21.69 & 46.57 \\
 & \ours   & 79.17 & 68.26 & 34.46 & 52.50 \\
 & \ours (w/ TTS) & 83.33 & \textbf{69.70} & \textbf{36.99} & \textbf{54.55} \\
\midrule
\multirow{4}{*}{Qwen3-32B} 
 & Direct GRPO & 86.46 & \textbf{80.10} & 21.82 & 53.73 \\
 & Direct DAPO & 86.46 & 77.08 & 26.18 & 54.08 \\
 & \ours   & 85.42 & 76.57 & 42.12 & 59.90 \\
 & \ours (w/ TTS) & \textbf{87.50} & 78.34 & \textbf{45.30} & \textbf{62.13} \\
\midrule
\multirow{4}{*}{LLaMA-3.3-70B} 
 & Direct GRPO & 79.17 & 63.48 & 25.23 & 46.81 \\
 & Direct DAPO & 75.83 & 63.94 & 22.61 & 45.65 \\
 & \ours   & 82.29 & 70.53 & 36.97 & 55.04 \\
 & \ours (w/ TTS) & \textbf{83.33} & \textbf{73.80} & \textbf{39.16} & \textbf{57.27} \\
\bottomrule
\end{tabular}

\caption{Comparison of our training framework against baselines on Fact Checking (FC), Numerical Reasoning (NR), and Data Analysis (DA) tasks. Table-R1 and Table-LLM results are directly extracted from \citet{wu2025table} (CoT prompts). For single-pass inference, all models are evaluated with temperature $0$, while parallel scaling is conducted with temperature $1$. The largest values within each model block are bolded.}
\label{tab:ours_training}
\end{table*}

For Table-LLM and Table-R1 in \tabref{tab:ours_training}, we report results as provided by \citet{wu2025table, wu2025tablebench}. 
%
From \tabref{tab:ours_training}, we see that large proprietary models such as o4-mini-high and DeepSeek-R1 set strong baselines, but our framework helps open-source small LLMs closes the gap. 
Notably, after applying our framework to Qwen3-32B, it raises its weighted average to 59.9\%, essentially matching o3-mini (59.9\%). 
Moreover, with test-time scaling (TTS), Qwen3-32B further improves to 62.13\%, surpassing o4-mini-high (61.69\%), the strongest baseline among proprietary models. 
For more details about TTS, including both parallel and sequential strategies, we will introduce it in \secref{subsec:ablation_tts}.

Compared with GRPO and DAPO, our framework consistently provides additional gains. 
On Qwen3-8B, GRPO and DAPO reach 53.7\% and 52.6\%, respectively, while our method improves this to 57.4\%.
When compared to prior table-specific models, our framework also shows clear advantages. 
For Qwen3-8B, \ours reaches 57.4\%, surpassing Table-R1 (49.3\%) and Table-LLM (35.9\%) by a large margin. 
For LLaMA-3.1-8B, our method boosts performance to 46.7\%, making it far ahead of Table-R1 (42.8\%) and Table-LLM (30.8\%). 
Moreover, we observe similar improvements on the out-of-domain FinQA dataset, where, for example, Qwen3-32B improves from 46.4\% with GRPO to 58.1\% with our framework, confirming that the benefits extend beyond the in-domain setting. 

Overall, \ours consistently achieves the best results across different model sizes, outperforming GRPO, table-specific baselines, and cutting-edge closed-source models.

\section{Ablation Studies and Analysis}
\label{sec:ablation_studies}

%

\subsection{Ablation on Sequential Training}
In this section, we analyze how each stage of our sequential training recipe contributes to the final performance.
As the training recipe is applied stage by stage, we evaluate performance after each stage to quantify the incremental gains from optimizing the planning, coding, and answering agents.
%

\begin{figure}[!tb]
    \centering
    \includegraphics[width=\linewidth]{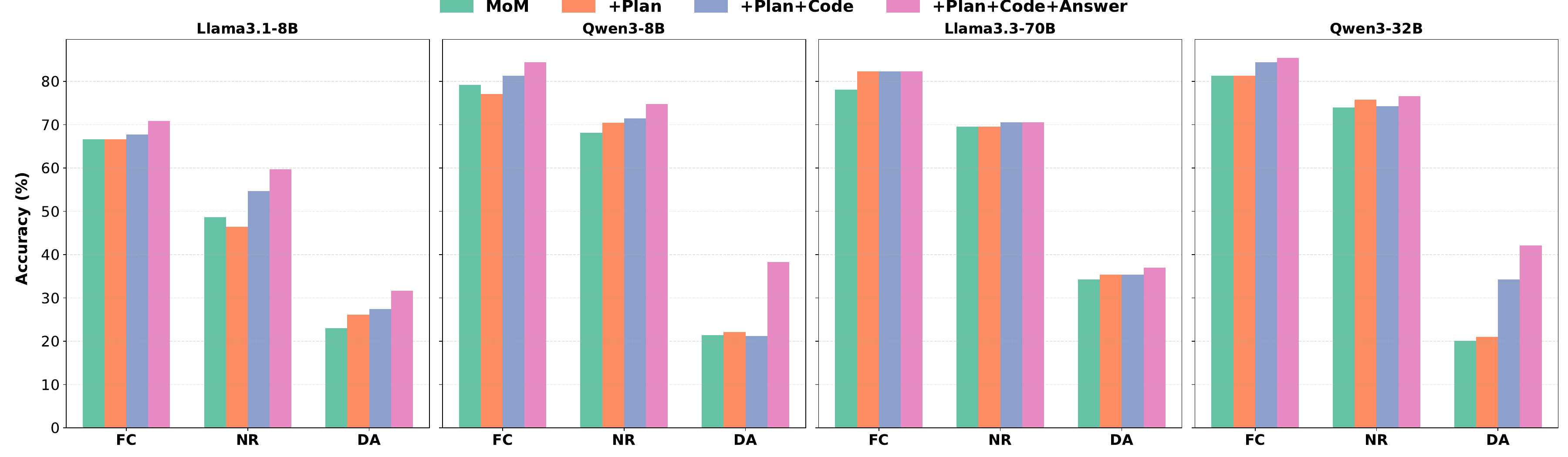}
    \caption{Ablation studies across different models. Each subplot shows performance on Fact Checking (FC), Numerical Reasoning (NR), and Data Analysis (DA) under sequential training stages: \ours w/o training, +Plan agent training, +Plan+Code agent training, and +Plan+Code+Answer agent training.}
    \label{fig:ablation_plot}
\end{figure}

As shown in \figref{fig:ablation_plot}, we see that sequential training steadily improves performance by training each agent. 
On LLaMA-3.1-8B, fact-checking accuracy remains flat at 66.7\% with \ours alone, but increases to 67.7\% after training the coding agent and further to 70.8\% after training the answering agent; numerical reasoning also improves step by step, from 46.5\% to 54.7\% to 59.7\%. 
A similar pattern holds for Qwen3-8B, where fact-checking grows from 77.1\% to 81.3\% to 84.4\%. 
Notably, bigger models also benefit from this incremental training: on LLaMA-3.3-70B, data analysis improves sequentially from 34.2\% to 35.4\% to 36.9\%, and on Qwen3-32B, data analysis rises from 20.1\% to 34.3\% and finally to 42.1\%. 
In 8 out of 12 model–task combinations, sequentially training the agents leads to monotonic improvements.

Among all tasks, training the answering agent tends to bring the most visible gains, especially in data analysis (e.g., Qwen3-8B improves from 21.2\% to 38.3\% and Qwen3-32B from 34.3\% to 42.1\%). 
This is expected because the answering agent directly aligns the final outputs with ground-truth answers, correcting residual errors even when planning and coding are accurate. 
In contrast, training the coding agent primarily improves execution validity and intermediate reasoning, yielding more moderate gains. 
This may also stem from the underlying coding capabilities of different LLMs, however, exploration in this direction is beyond the scope of this work.
Taken together, these results demonstrate that while each agent contributes in different ways, the full sequence of planning, coding, and answering training delivers the largest overall improvement.

\subsection{Test-time Scaling of \ours}
\label{subsec:ablation_tts}

%

In this section, we explore how our proposed \ours framework enables different forms of TTS.
%
Specifically, we investigate two TTS strategies: 
(1) \textit{parallel scaling}, where the workflow is executed multiple times and aggregated with self-consistency~\cite{wang2022self}; 
and (2) \textit{sequential scaling}, where the coding agent regenerates code until execution succeeds, adaptively increasing tokens per inference. 
This allows us to understand how \ours trades additional inference-time compute for better performance.

For parallel scaling, we repeat the planning agent $8$ times for early branching (temperature 1.0), while running the coding and answering agents once, resulting in $8$ diverse traces that are aggregated by self-consistency~\citep{wang2022self}.
This is applied to both \ours with our training framework and to direct inference as a baseline. 
For sequential scaling, we focus on \ours with training: when the coding agent produces code that fails to execute, we regenerate by providing the historical code and error message until success. 
%
%
Results are reported in \tabref{tab:test_time_scaling}.

\begin{table}[t]
\setlength{\tabcolsep}{16pt}
\centering
\small
\begin{tabular}{l l cccc}
\toprule
Model & Setting & FC & NR & DA & Avg \\
\midrule
\multirow{5}{*}{Qwen3-8B} 
 & Base (Single) & 78.12 & 53.90 & 24.11 & 41.98 \\
 & Base (Para.) & 79.17 & 51.89 & 26.06 & 41.94 \\
\cmidrule(l){2-6}
 & \ours{} (Single) & 84.38 & 74.75 & 38.29 & 57.44 \\
 & \ours{} (Seq.) & 84.38 & 76.32 & 40.94 & 59.20 \\
 & \ours{} (Para.) & \textbf{86.46} & \textbf{77.83} & \textbf{41.59} & \textbf{60.35} \\
\midrule
\multirow{5}{*}{Qwen3-32B} 
 & Base (Single) & 81.25 & 72.54 & 20.06 & 49.02 \\
 & Base (Para.) & 82.29 & 72.29 & 22.64 & 50.12 \\
\cmidrule(l){2-6}
 & \ours{} (Single) & 85.42 & 76.57 & 42.12 & 59.90 \\
 & \ours{} (Seq.) & 85.42 & 76.32 & 42.90 & 60.04 \\
 & \ours{} (Para.) & \textbf{87.50} & \textbf{78.34} & \textbf{45.30} & \textbf{62.13} \\
\midrule
\multirow{5}{*}{LLaMA-3.1-8B} 
 & Base (Single) & 54.17 & 7.30 & 16.20 & 15.42 \\
 & Base (Para.) & 54.79 & 10.33 & 13.57 & 15.84 \\
\cmidrule(l){2-6}
 & \ours{} (Single) & 70.83 & 59.70 & 31.69 & 46.72 \\
 & \ours{} (Seq.) & 78.12 & 62.47 & 49.47 & 49.27 \\
 & \ours{} (Para.) & \textbf{81.25} & \textbf{67.51} & \textbf{36.80} & \textbf{53.31} \\
\midrule
\multirow{5}{*}{Gemma3-27B} 
 & Base (Single) & 79.17 & 54.66 & 24.73 & 42.67 \\
 & Base (Para.) & 79.17 & 52.39 & 23.76 & 41.26 \\
\cmidrule(l){2-6}
 & \ours{} (Single) & 79.17 & 68.26 & 34.46 & 52.50 \\
 & \ours{} (Seq.) & 79.17 & 68.51 & 35.99 & 53.23 \\
 & \ours{} (Para.) & \textbf{83.33} & \textbf{69.70} & \textbf{36.99} & \textbf{54.55} \\
\bottomrule
\end{tabular}
\caption{
    Test-time scaling analysis results on Fact Checking (FC), Numerical Reasoning (NR), and Data Analysis (DA) tasks. 
    ``Base'' denotes direct model inference with base LLMs. 
    We report performance under single-pass (1$\times$), parallel scaling (8$\times$), and sequential scaling (repeat until executable code).
}
\label{tab:test_time_scaling}
\end{table}
%



From \tabref{tab:test_time_scaling}, we find that applying TTS to \ours consistently improves performance, boosting open-source models beyond o4-mini-high. 
Notably, Qwen3-32B with \ours under parallel scaling achieves 62.13\%, surpassing o4-mini-high's 61.7\%, a substantial margin given the competitiveness of proprietary systems.
On average, sequential scaling improves performance by +1.30\%, and parallel scaling by +3.65\%, with all models improving under both strategies. 
By contrast, model-only inference gains little from TTS, yielding only marginal gain and improving only half of the evaluated cases.
%
%

\section{Conclusion}
\label{sec:conclusion}

In this work, we introduced \ours{}, a novel agent workflow for table reasoning, together with a compatible sequential training framework that leverages MCTS-style data generation and GRPO optimization with designed reward functions. Through extensive experiments, we demonstrated that \ours{} achieves substantial improvements over baseline methods, surpasses strong closed-source systems under test-time scaling, and consistently benefits from sequential training as shown in our ablation studies. These results highlight the value of structured, multi-agent reasoning. 


For future work, we expect that the proposed MCTS-based strategy for extracting intermediate supervision can be extended to more flexible agent workflows beyond table reasoning. This framework could support dynamic collaboration among multiple agents specialized in diverse tasks such as planning, coding, or decision-making, paving the way for broader applications of multi-agent learning.








\clearpage
\newpage
\bibliographystyle{assets/plainnat}
\bibliography{custom}

\clearpage
\newpage
\beginappendix
\appendix
\section*{Appendix}

\section{Prompt Details}
\label{sec:prompts}

We present the prompt details for constructing our \ours{} agent workflow in Table \ref{tab:prompt}.

\begin{table}[h]
\centering
\small
\begin{tabular}{p{0.99\linewidth}}
\toprule
\textbf{Planning Agent (System)} \\
\midrule
You are an expert data analyst specializing in table analysis. Your task is to create a clear, step-by-step analysis plan to guide another LLM to generate pandas code to answer questions about tabular data. Guidelines: break down into logical steps, simple focused operations, actionable with pandas, no visualization, be specific, each step results in a new table closer to the answer, etc. Output enclosed in \texttt{<plan>...</plan>}. \\
\midrule
\textbf{Planning Agent (User)} \\
\midrule
**Question and Input Table** \{instruction\}. Your task is to create a detailed analysis plan with at most 4 steps. Each step should describe what to extract, what to do, and why it helps answer the question. Be specific about metrics, columns, or operations. \\
\midrule
\textbf{Coding Agent (System)} \\
\midrule
You are a coding agent using pandas. Write full Python code (with imports, table loading, and execution) to follow the given plan. No plotting. At the end, print (1) the answer and (2) a short summary. Output must be wrapped in \texttt{``python ...''}. \\
\midrule
\textbf{Coding Agent (User)} \\
\midrule
**Input table and question**: \{instruction\}. The planner already made a plan: \{plan\}. Write code to execute it. Handle data types and empty cells carefully. \\
\midrule
\textbf{Answering Agent (System)} \\
\midrule
You are a data scientist specializing in table analysis. Answer questions with accuracy, critical thinking, and domain knowledge. Apply causal reasoning, avoid trivial correlations, be skeptical of artifacts, and follow requested answer format exactly. \\
\midrule
\textbf{Answering Agent (User)} \\
\midrule
**Answer format, question, and table**: \{instruction\}. **Plan**: \{plan\}. **Code output**: \{code\_output\}. If errors occur in code output, omit it. Otherwise, verify correctness and provide the final answer. \\
\bottomrule
\end{tabular}
\caption{System and user prompt templates for planning, coding, and answering agents in \ours{}.}
\label{tab:prompt}
\end{table}

\begin{table*}[t]
\setlength{\tabcolsep}{25pt}
\centering
\small
\begin{tabular}{l l c}
\toprule
Model & Method & FinQA (\%) \\
\midrule
\multirow{3}{*}{Qwen3-8B}
 & Direct GRPO & 44.57 \\
 & \ours{} (w/o finetune) & 54.63 \\
 & \ours{}  & \textbf{55.25} \\
\midrule
\multirow{3}{*}{LLaMA-3.1-8B}
 & Direct GRPO & 27.18 \\
 & \ours{} (w/o finetune) & 24.54 \\
 & \ours{}  & \textbf{33.80} \\
\midrule
\multirow{3}{*}{Gemma3-27B}
 & Direct GRPO & 50.93 \\
 & \ours{} (w/o finetune) & 48.46 \\
 & \ours{}  & \textbf{51.89} \\
\midrule
\multirow{3}{*}{Qwen3-32B}
 & Direct GRPO & 46.43 \\
 & \ours{} (w/o finetune) & 56.57 \\
 & \ours{}  & \textbf{58.08} \\
\midrule
\multirow{3}{*}{LLaMA-3.3-70B}
 & Direct GRPO & 50.57 \\
 & \ours{} (w/o finetune) & 48.10 \\
 & \ours{}  & \textbf{51.89} \\
\bottomrule
\end{tabular}
\caption{Out-of-domain results on FinQA. We compare a direct GRPO baseline with our \ours{} pipeline (without training) and \ours{} with our training framework . Best result per model is bolded. All runs use a single inference pass with temperature $0$.}
\label{tab:ours_training_finqa}
\end{table*}

\section{Hyperparameter Details}
\label{sec:hyper}

The numbers of sampled plans ($\alpha$), codes ($\beta$), and answers ($\gamma$) during MCTS rollouts differ by agent, as summarized in \tabref{tab:mcts_hparams}. For GRPO optimization, we use a group size of $G=8$ (as defined in Section \ref{section:grpo}). The learning rate is set to $1\times10^{-6}$, with a global batch size of $256$ and a rollout temperature of $1.0$. Each experiment is trained for approximately $100$ update steps.

\begin{table}[!tb]
\centering
\small
\begin{tabular}{lccc}
\toprule
Agent & $\alpha$ (Plans) & $\beta$ (Codes) & $\gamma$ (Answers) \\
\midrule
Planning   & 8 & 4 & 1 \\
Coding     & 1 & 8 & 1 \\
Answering  & 1 & 4 & 8 \\
\bottomrule
\end{tabular}
\caption{MCTS rollout hyperparameters for different agents, where $\alpha$, $\beta$, and $\gamma$ denote the number of sampled plans, codes, and answers, respectively.}
\label{tab:mcts_hparams}
\end{table}

\section{Data Processing}
\label{sec:data_processing}
The TableInstruct \cite{shao2024deepseekmath} dataset was constructed from multiple existing table datasets, including FinQA, TabFact, etc. Both questions and answers were constructed and validated using GPT-4-based agents. The answers were constructed with four approaches: DP (direct prompting), TCoT (textual chain-of-thought), SCoT (symbolic chain-of-thought) and PoT (program-of-thoughts), thus for each question, there are four answers in the dataset. 

Since the questions/answers were only generated by LLM and not fully manually verified, we first investigated the dataset for correctness, and found two types of errors: 1) Answers generated with different approaches (DP / TCoT / SCoT / PoT) for the same question do not agree with each other. This appears mainly in Fact Checking and Numerical Reasoning questions, since there should only be one correct answer. 2) Wrong answers for Data Analysis categories. We identified several wrong answers in the dataset, such as mixing causality with correlation (e.g. the question is asking about causality, but the answer only provided correlation) and wrong value for trend forecasting. 

\paragraph{Data cleaning for Fact Checking and Numerical Reasoning questions.} We first applied rule-based filtering for Fact Checking and Numerical Reasoning data: for each question, the answer is correct, only the answers generated by all four methods are the same (DP / TCoT / SCoT / PoT). We normalize the format of the answers, such as numerical precision, before comparing them. 

For the remaining questions, we used a coding agent to derive the answer. The coding agent produces runnable code to solve the problem and then answers the question from the code output. We used the Claude Sonnet 4 and Google Gemini 2.5 Pro models as the agent backend to generate two sets of answers for the same question. 

We then compared the generated answers with the DP answer from the original dataset: if the DP answer matches one of the generated answers (either from Gemini or Claude agents output), we consider the DP answer correct. This resulted in 556 questions whose answers need to be updated. Next we check if the answer from two coding agents from Claude / Gemini agree with each other, and consider the generated answers as the correct one. The last 177 questions with different generated answers were reviewed by the authors and their final answers manually created.

\paragraph{Data cleaning for Data Analysis questions.} Since the Data Analysis answers are provided in natural language, it is hard to use exact match to filter out wrong answers. We thus used LLM-as-judge design to extract the correct answers. We first applied a coding agent to generate a new answer for the question, then feeding both the original answer, the code output and the generated answer to another LLM to decide the final correct answer. We used Gemini Pro 2.5 as the LLM backend.

We then manually inspect a set of randomly sampled rows to ensure the data quality. This method is not 100\% correct since the judge can make errors, yet we expect higher quality of the updated dataset, as of 1) the LLM base model has higher performance than the GPT-4 model used in the original paper published in 2024; 2) the coding agent providing a grounded reasoning process validated by external (python) tools. In the end, 1463 Data Analysis questions' answers were updated by the LLM.

\section{Coding Agent Reward Design Details}
\label{sec:reward}

The coding agent reward consists of four components: format, execution, operation, and output rewards. For completeness, we provide further details here.

\paragraph{Format Reward ($r^{\text{fmt}}$).} 
Ensures that generated code is wrapped within proper Python markdown, enabling consistent parsing and execution during evaluation. 

\paragraph{Execution Reward ($r^{\text{exec}}$).} 
Assigned as $1$ if the generated code executes successfully without runtime errors. This enforces syntactic and runtime validity. 

\paragraph{Operation Reward ($r^{\text{op}}$).} 
We extract high-level Pandas operations (e.g., \texttt{groupby}, \texttt{merge}, \texttt{pivot}) from both the generated code $c_i$ and gold code $c^\star$, normalize their order, and compute an F1 score over the operation set. This encourages structural similarity without requiring exact token-level matches. 

\paragraph{Output Reward ($r^{\text{out}}$).} 
Unlike $r^{\text{exec}}$ (which checks only validity), this reward measures the \emph{semantic correctness} of the generated code output relative to the gold-standard $c^\star$. We compute $r^{\text{out}}$ as the BLEU score \cite{papineni2002bleu} between the predicted output string and the reference output. This design allows partial credit for outputs that are semantically close but not exact matches, and stabilizes training compared to a strict binary correctness reward.

Thus, the final coding agent reward balances surface validity (formatting, execution) with structural similarity and end-to-end correctness.

\clearpage
\section{Supplementary Results}

\subsection{FinQA Evaluation}
\label{sec:finqa_result}

To better demonstrate the generalization ability of our training framework, we also evaluate the trained agents with \ours{} on the FinQA dataset as an out-of-domain benchmark. The results are presented in Table~\ref{tab:ours_training_finqa}.

\subsection{Reward Curve Plot}
\label{sec:plot}

\begin{figure*}[t]
    \centering
    \begin{subfigure}{0.32\linewidth}
        \centering
        \includegraphics[width=\linewidth]{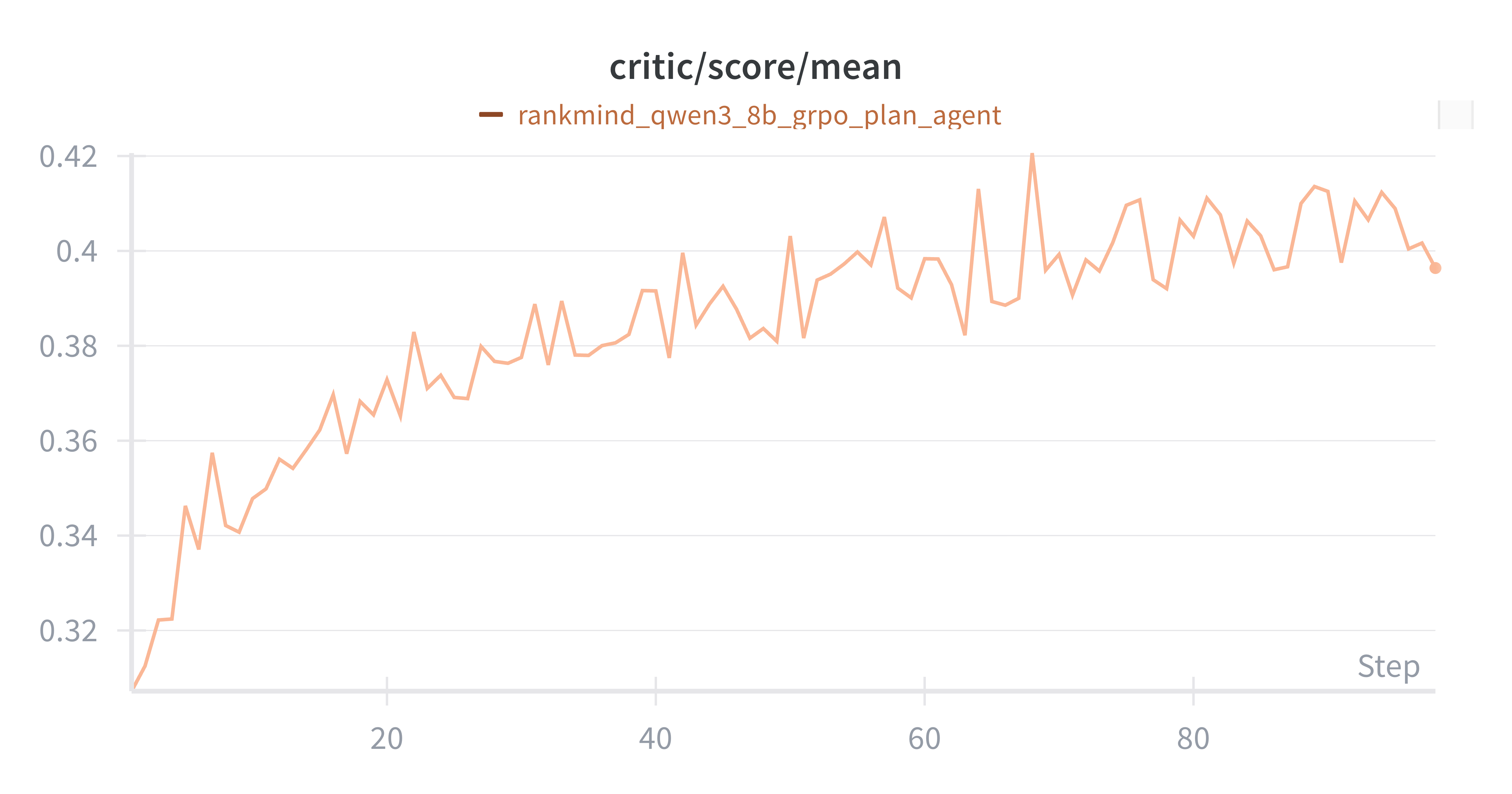}
        \caption{Planning Agent}
        \label{fig:reward_qwen8b_plan}
    \end{subfigure}
    \hfill
    \begin{subfigure}{0.32\linewidth}
        \centering
        \includegraphics[width=\linewidth]{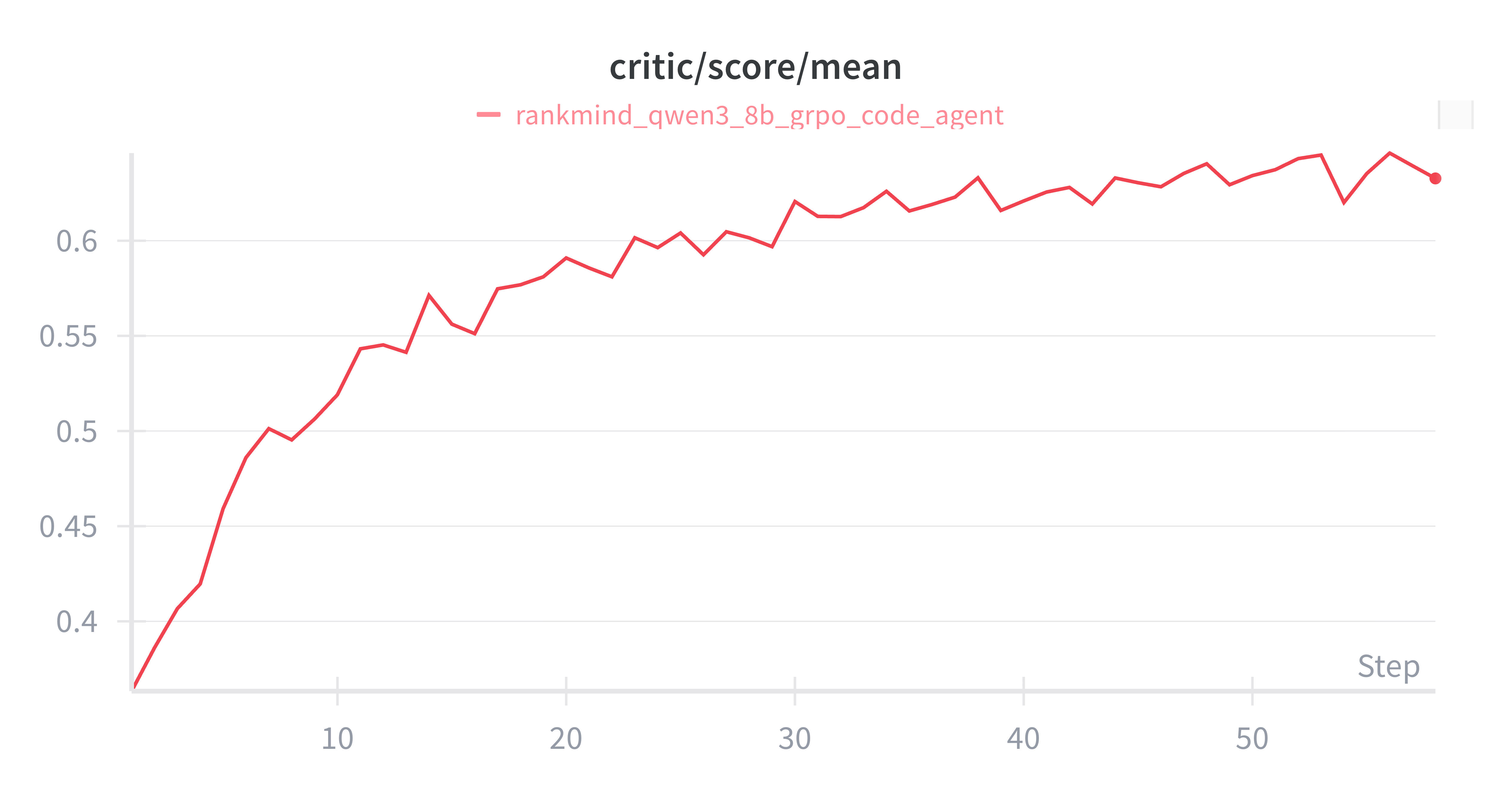}
        \caption{Coding Agent}
        \label{fig:reward_qwen8b_code}
    \end{subfigure}
    \hfill
    \begin{subfigure}{0.32\linewidth}
        \centering
        \includegraphics[width=\linewidth]{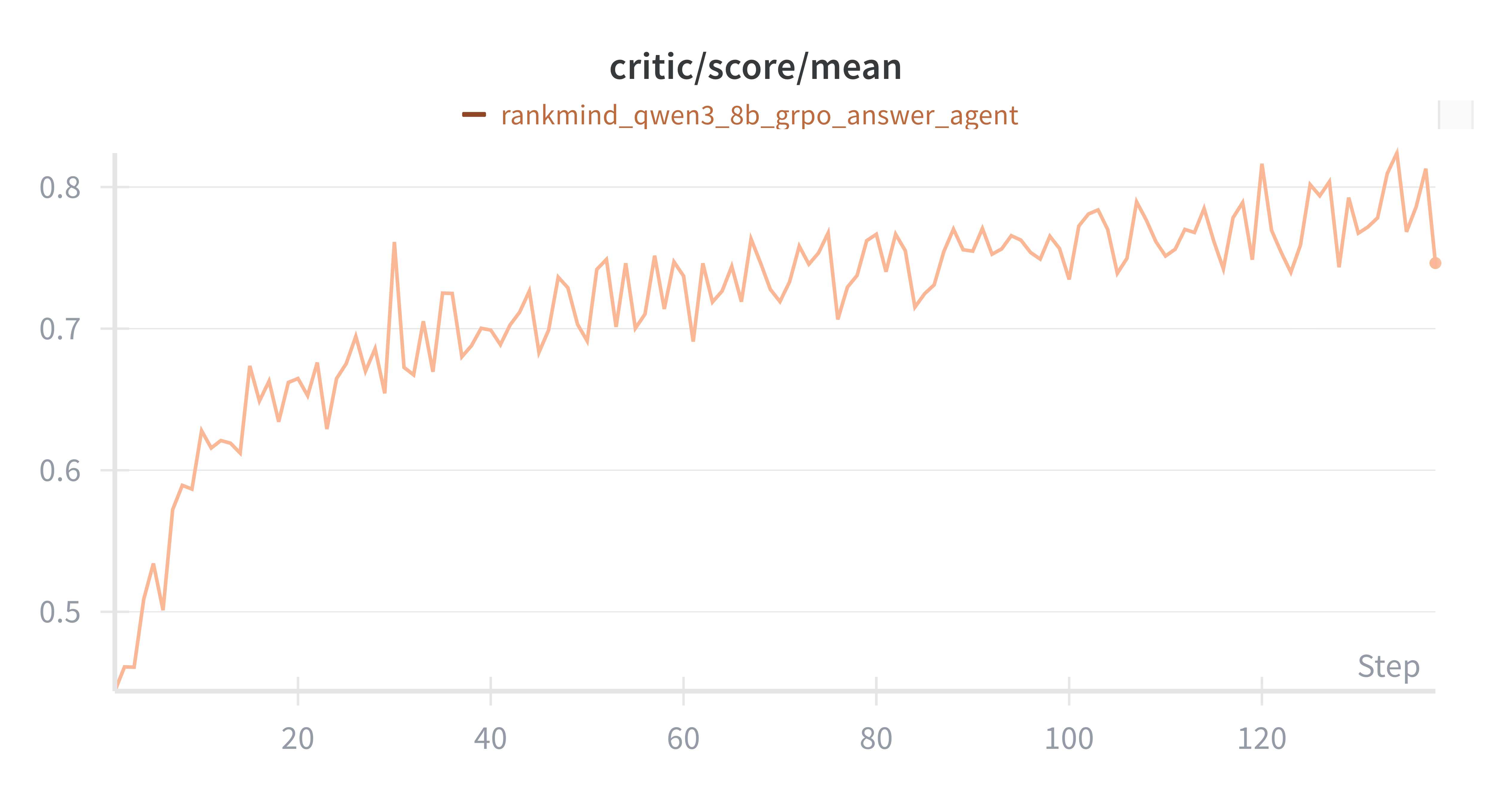}
        \caption{Answering Agent}
        \label{fig:reward_qwen8b_answer}
    \end{subfigure}

    \caption{Reward trajectories of the planning, coding, and answering agents on Qwen3-8B. Each plot illustrates how our designed reward functions guide learning over training.}
    \label{fig:reward_plot_qwen8b}
\end{figure*}

To better understand how each component contributes during training, we visualize the reward trajectories of the planning, coding, and answering agents on Qwen3-8B in Figure~\ref{fig:reward_plot_qwen8b}.

\end{document}